# PRM path smoothening by circular arc fillet method for mobile robot navigation*

Meral Kılıçarslan Ouach[1]**, Tolga Eren[2]

[1,2]Department of Electrical-Electronical Engineering, Kırıkkale University, 71451, Kırıkkale Turkey

**Abstract**

Motion planning and navigation, especially for mobile robots operating in complex navigational environments, has been a central problem since robotics started. A heuristic way to address it is the construction of a graph-based representation (a path) capturing the connectivity of the configuration space. Probabilistic Roadmap is a commonly used method by the robotics community to build a path for navigational mobile robot path planning. In this study, path smoothening by arc fillets is proposed for mobile robot path planning after obtaining the path from PRM in the presence of the obstacle. The proposed method runs in two steps; the first one is generating the shortest path between the initial state to one of the goal states in the obstacle presence environment, wherein the PRM is used to construct a straight-lined path by connecting the intermediate nodes. The second step is smoothening every corner caused by node presence. Smoothening the corners with arc fillets ensures smooth turns for the mobile robots. The suggested method has been simulated and tested with different PRM features. Experiment results show that the constructed path is not just providing smooth turning; it is also shorter and quicker to finish for a robot while avoiding obstacles.

## 1. Introduction

Setting up a collision-free path and motion planning for autonomous vehicles to move from an initial state to a final destination, especially for obstacle hindered environments, has been challenging work in robotics over the last decade [1]–[3]. Even though the latest robots may possess substantial differences in sense, size, actuation, workspace, application, etc., the problem of navigating through a complex environment is compound and crucial in almost all robotics applications. This problem also appears relevant to other domains such as autonomous exploration, computational biology, agriculture, search and rescue, etc. Considering the affecting factors on path planning (PP) like different kinds of robots and navigational environments, multiple robots, and dynamic or static obstacles, finding the shortest path with the highest degree of smoothness prevents collision with obstacles and other robots is still a challenging problem.

The goal of PP for mobile robots is to find a collision-free path from the starting point to the target point and optimize it based on specific methods [4], [5]. Thus, determining an optimal path is difficult within environments containing many obstacles. As a result, depending on the complexity of the environment, this problem is categorized as NP-hard or NP-complete [6], [7]. Therefore, PP studies generally specify the assumed environment type, known/unknown environments based on a pre-identified map. Nevertheless, the environment can be dynamic or static depending on whether the obstacle position changes when the mobile robot moves; however, a reliable map is essential for navigation without which robots cannot achieve their goals.

Another critical feature of robot PP is path smoothening. Although usually, the simplified PP solutions compute piecewise linear paths that neglect robot restrictions, it can sometimes be possible

---

that the planned path may not be practical for a particular robot because the robot kinematic or other restrictions are not taking account of or the path may include sharp corners in which the robot cannot perform successfully. Thus, smoothening the path is very much important in these conditions. The path smoothening supports synchronization of robot acceleration and velocity with robot kinematics in practical applications such as: in case of multiple robot navigation, not only the navigation path but also the firm control over the robot acceleration and velocity are important points, or for the robot manipulator's path planning, the smoothening is necessary for effectively controlling the acceleration and velocity at the path corners so that the resulting accelerations or velocities can be maximized within their limitations, etc. Therefore, there are many reasons for path smoothening.

| **Nomenclature** | | MD | Morphological Dilation |
|---|---|---|---|
| Abbreviation, | Definition | VD | Voronoi Diagram |
| $\mathcal{G}$ | Graph | RRT | Rapidly-exploring Random Trees |
| $\mathcal{V}$ | Vertex | APF | Artificial Potential Field |
| $\mathcal{E}$ | Edge | GA | Genetic Algorithm |
| PRM | Probabilistic Roadmap | ABC | Artificial Bee Colony |
| CS | Configuration Space | EPA | Evolutionary Programming Algorithm |
| dist | Distance | QPI | Quadratic Polynomial Interpolation |
| PP | Path Planning | SVM | Support Vector Machine |
| PSO | Particle Swarm Optimization | Eqn. | Equation |

However, studies in the literature mentioned in Section 3 show that considerably less effort was made to smooth the path or have time complexity. For example, it is mentioned in [8] that a circular arc is usually configured to replace the joints of the path segments so that a smooth path can be performed. On the other hand, Mohanta and Keshari in [9] state that drawing a circular arc can be problematic if path joints are very close to the obstacles, and determining the appropriate radius of the circular arcs is itself difficult. However, utilizing a modified geometric approach proposed in this paper, determining a suitable circular arc radius becomes easy to compute and perform. Moreover, path safety is ensured by inflating the obstacles before applying PRM.

In this paper, new modified circular arcs are introduced to replace path segment joints so that optimal smooth path can be performed and smooth turning of the robot movement. This method uses the advantage of Probabilistic Roadmap (PRM), one of the most popular applications for computing navigational paths. The presented arc fillet method is a powerful study, especially for smoothening the robot navigational path, which can be counted as the novelty of this study. The success of the proposed approach is demonstrated through many simulations, and its superiorities are discussed thoroughly.

The primary components of the framework of the proposed method include workspace preparation, path computation, and path smoothing. The framework is presented in Fig. 1, and the general algorithm for the PRM Arc Fillet method is shown in a flowchart in Fig. 14.

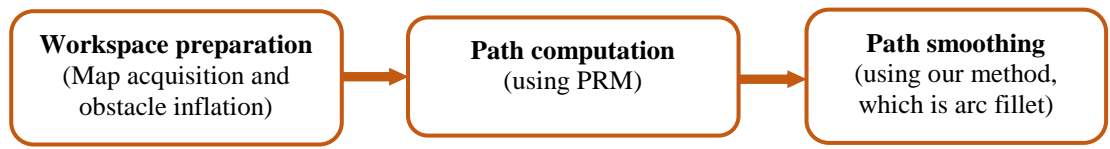

**Fig. 1.** Workflow diagram of the proposed method

In Section 0 of the study, a comprehensive definition of PRM has been given. Section 3 has been expanded by ensuring a broad perspective to the studies in the literature due to the suitability of the problem to PP methods. Section 4 demonstrates the calculations of the optimal radius of circular arcs and optimal intersection points of arcs with the path edges. Path smoothing simulations at different PRM features are shown, and comparisons are made in Section 5. Finally, the obtained results and discussions are compiled in Section 6, and Section 7 concludes the outcomes, interests, and future scope of the research.

## 2. Mathematical Model

### 2.1 Probabilistic Roadmap (PRM)

Classic PRM planner is a motion planning algorithm that makes sample points randomly defined as location nodes at the robot's configuration space and connects these points to construct the roadmap graph that captures connectivity of the collision-free subset of the configuration space (CS) [10]. It is highly appreciated to determine a pre-planned mobile robot path that solves the problem of calculating a path between the starting point of a robot to the goal point(s) while avoiding collisions with obstacles. PRM is combined by graph search algorithm (breadth-first search) to make graphs at CS during the construction phase and Dijkstra algorithm to obtain the shortest path between starting and goal points during the query phase.

***Dijkstra Algorithm:*** Dijkstra's algorithm finds the shortest path between the start point and the endpoint among the location nodes in a workspace [11]. It is based on Euclidian distance as a measurement when the distance between nodes needs to be calculated. Dijkstra's algorithm, to start with, marks the distance from the starting point to every other node. Initially, the starting point, which is the source, is labeled as zero distance, while every other node is marked as infinity (see Algorithm 1). By exploring every node, their infinite value is changed to a constant number. Dijkstra's algorithm repeats a two-step process; the first step is updated estimates, and the second is the next nodes chosen to be explored. From the starting point, the shortest distance is selected among the nodes connected to the starting point, and then this point is labeled as the second point. This step is repeated until reaching the endpoint to find the shortest path.

---

**Algorithm 1**: Operation of Dijkstra

**Input**: $\mathcal{N}$ random samples from *CS*
**Output**: A trajectory $\mathcal{G} = (\mathcal{V}, \mathcal{E})$ in *CS*
1:   $\mathcal{V} \leftarrow \{SampleFree_i\}_{i=1,\ldots,\mathcal{N}}$
2:   $\mathcal{E} \leftarrow \emptyset$
3:   dist[source]←0
4:   **for each** $v \in \mathcal{G}$ **do**
5:       dist[$v$] ← infinity
6:       prev[$v$] ← null
7:       add $v$ to $\mathcal{V}$
8:       **while** $\mathcal{G}$ is not empty, **then**
9:           $u \leftarrow v$ with min dist[$u$]
10:         remove $u$ from $\mathcal{V}$
11:       **end while**
12:   **end for**
13:   **return** $\mathcal{G}$

---

The goal of PRM is to calculate an optimal collision-free trajectory. The Standard algorithm for construction PRM is reviewed in Algorithm 2. Some formal definitions used in the description of the algorithm are as follows; let *CS* be the configuration space where $d \in \mathbb{N}$ is its dimension. The PRM establishes a roadmap stated as a graph $\mathcal{G} = (\mathcal{V}, \mathcal{E})$ whose vertices are selections from *CS* and the edges are collision-free routes between vertices, as shown in Fig. 2. Then the PRM launches the vertex set with $\mathcal{N}$ random samples from *CS* and tries to connect the nearest nodes. If no path is found, then nodes at *CS* must be increased at the construction phase.

---

**Algorithm 2**: Operation of PRM

**Input**: $\mathcal{N}$ random samples from *CS*
**Output**: A trajectory $\mathcal{G} = (\mathcal{V}, \mathcal{E})$ in *CS*
1:   $\mathcal{V} \leftarrow \{SampleFree_i\}_{i=1,\ldots,\mathcal{N}}$

```
2:   ℰ ← ∅
3:   for each 𝓋 ∈ 𝒱 do
4:       𝒰 ← Apply_Dijkstra(𝒢, 𝓋)
5:       for each 𝓊 ∈ 𝒰 do
6:           if CollisionCheck(𝓋, 𝓊) then
7:               ℰ ← ℰ ∪ {(𝓋, 𝓊), (𝓊, 𝓋)}
8:           end if
9:       end for
10:  end for
11:  if 𝒢 is not computed, then
12:      return no path is found
13:  else
14:      return 𝒢
15:  end if
```

Mobile robot path planning by PRM is required in these steps, i.e., import or create map depiction of the obstacle present environment, plan an obstacle-free path from a start and goal points at CS, finally follow the planned path from starting point to endpoint. In this study, maps are imported from the MATLAB library as simple and complex maps. The PRM complex map was inflated as obstacles to have a safe environment for the robot during the navigation. Map inflation extends each occupied position to a specified amount in the input by radius selected in meters as obstacles. Radius is rounded up a similar neighboring cell based on the resolution of the map. This map inflation increases the size of the occupied positions in the map. After path construction finishes, the inflated map returns to its original. This method is useful when the path is reshaped after PRM output.

Mobile robot path planning and executing the robot movements in an obstacle-free path using PRM involves various steps. The distinct representation of the mobile robot surroundings is the first important step. An area map obtained from images or manually can also be created using range sensor data from the robot and an occupancy grid that obstacle presence locations. In the second step, the PRM path planner calculates the connectivity between different map areas, extracting the obstacle-free path using the various CS nodes. If the robot movement project can work successfully in the simulation environment, they can implement it in real-life experiments.

As required, the probabilistic roadmap method is related; it is formed by two phases, i.e., (1) construction phase and (ii) query phase. In the construction phase, a trajectory (graph) is built by corresponding to the motions made in CS. Firstly, random nodes are generated in CS. Then they are attached to some neighbors, typically neighbors less than some predetermined distance (see Fig. 3) or the *k*-nearest neighbors (see Fig. 2). Finally, connections and configurations are added to the graph until the roadmap is compact. The query phase operates; the start and goal positions are connected by a graph derived by Dijkstra's shortest path algorithm.

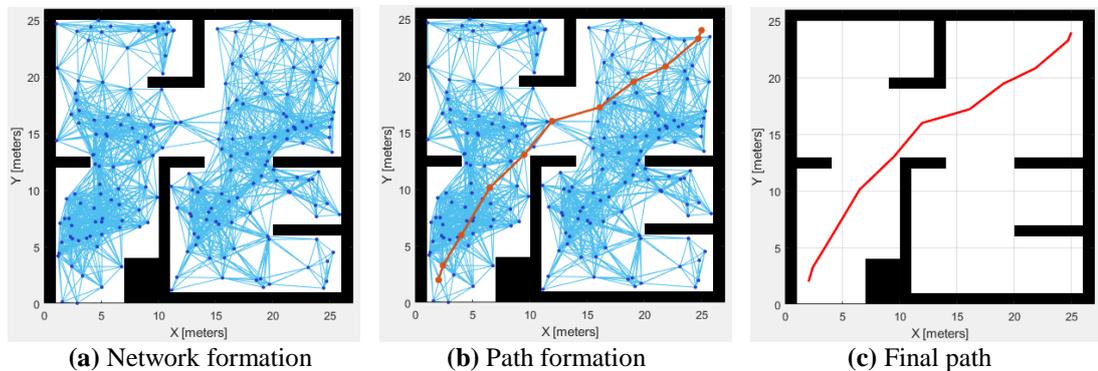

(a) Network formation       (b) Path formation       (c) Final path
**Fig. 2.** A Simple construction of PRM

A separate case study of PRM path formation that contains phases is demonstrated in Fig. 2. In network formation, nodes are joined using the breadth-first search graph traversal technique after a specified number of nodes are randomly created in the given map, as shown in Fig. 2a for the construction phase. Then Dijkstra's algorithm runs to determine of selection the shortest path as shown in Fig. 2b for query phase, and the final path can be seen better in Fig. 2c.

Node numbers can determine a straighter and shorter path at CS and a connection distance for the nodes at the path. Increasing node numbers brings computational complexity; hence the construction time of the path would be higher. At the same node numbers, different connection distance affects the shape of the path. When connection distance increases, the connection between nodes also increases during node formation; this gives more choices amongst nodes to form a path during path formation; hence the path would be straighter and shorter, but duration also increases. For example, at the PRM in Fig. 3, the connection distance is less than Fig. 2, but the path is straighter and shorter (Both figures have the same number of nodes, and these nodes are positioned at the same locations).

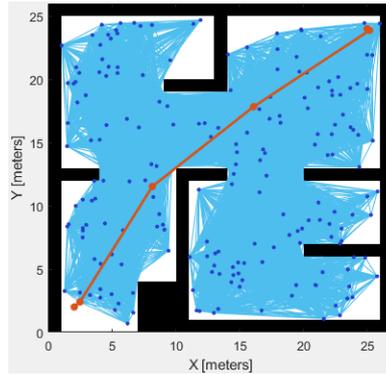

**Fig. 3** Effects of connection distance on PRM

A common version of the PRM is the lazy-collision PRM [12], [13] or its asymptotically optimal variant lazy-collision PRM*. In those versions of PRM, the ColisionCheck action in line 6 of Algorithm 1 is cut during the construction stage. Every time a path between two vertices is searched and detected in the graph, it is confirmed if at least one of its edges collides with an obstacle during the query stage. If one of the edges in the path is in a collision with an obstacle, that edge is eliminated from the graph, and a new path is constructed.

PRM benefits have been appreciated in various mobile robot path planning applications in 2D and 3D spaces, i.e., multiple robot path planning, pick and place robot end-effector path planning, etc. PRM is useful to predetermine obstacle-free the shortest path; this length can also be shortened by increasing the number of randomly specified nodes. If the number of nodes increases during the construction phase, then the length of path divisions reduces, in return, an increase in the number of intermediate nodes. As a result, an increase in the number of sharp edges is found at the path. The pre-planned route is nothing but the link of many straight lines connecting the start and goal positions. Straight-line connections cause sharp edges throughout the entire path. Whether the number of nodes is enhanced or not, PRM calculated path cannot prevent sharp edges.

*2.2 Circular Arc Fillet*

To be self-content, we briefly review an arc fillet definition. Arc fillet has good geometric properties and has been widely used in computer graphics and cad drawings applications in engineering and architectural projects. This algorithm detects the beginning angle and the angle subtended by the arc and the arc's direction. The lines' new beginning and ending points will be computed to join the arc smoothly [14]. One of the significant advantages of the arc fillet is that it can be applied to any line that intersects at any angle. Therefore, the arc fillet is a good tool for curve fitting, and it smooths the line segments by turning them into arcs.

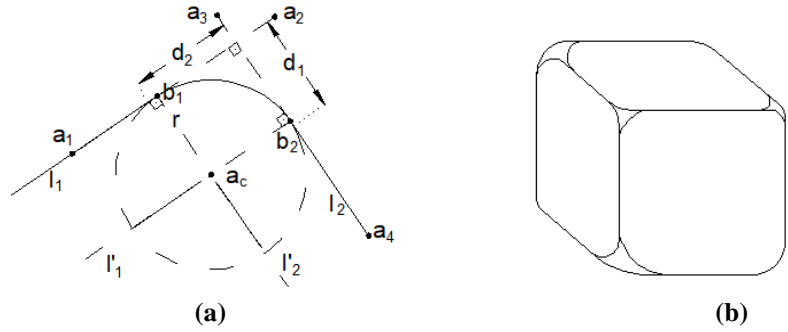

**Fig. 4.** Joining two lines **(a),** and three lines **(b)** with a circular arc fillet

Constructing a circular arc fillet that joins two lines, $l_1$ ($\overline{a_1 a_2}$) and $l_2$ ($\overline{a_3 a_4}$) (see Fig. 4a) can be performed step-by-step:

1) The line equations in the form $ax + by + c = 0$ need to be found for the lines that intersect. The center of the fixed arc, $a_c$, must lie at a distance $r$ of both lines.

Define the distance $d_1$ from $l_1$ to the midpoint of $l_2$ and $d_2$ from $l_2$ to the midpoint of $l_1$. The midpoints are used because, in practice, one point may be mutual to both lines. The signs of d1 and d2 determine on which sides of the particular lines the arc center ac resides (See Fig. 4a).

2) Find $l_1' \parallel l_1$ at $d_1$ and $l_2' \parallel l_2$ at $d_2$. The center of the expected arc $a_c$ lies at the intersection of $l_1'$ and $l_2'$.

3) Calculate the beginning and ending points, $b_1$ and $b_2$, on the arc.

4) Find the starting angle, $s$ (concerning the x-axis). Angle $s = \tan^{-1}(\overline{b_1 a_c})$. The two-argument arctangent is used to uniquely determine $\alpha$ in the range $0 \leq s < 2\pi$. Utilize the vector dot product to the directed line segments $b_1 a_c$ and $b_2 a_c$ to find the angle $\alpha$ subtended by the arc from $a_c$.

5) Use the vector cross product sign to define the direction to draw the arc $b_1 b_2$.

6) The line may extend or be clipped at points $b_1$ and $b_2$, so the endpoints of the nearest line point of intersection of $l_1$ and $l_2$ will coincide with the endpoints of the arc. The fillet will result from drawing a line from $a_1$ to $b_1$, the arc from $b_1$ to $b_2$, and the line from $b_2$ to $a_4$.

The arc fillets are well connected to form a smooth path planning for mobile robots in the path planning problem.

### 3. Related literature

The comprehensive research components of this study consist of mobile robot path planning using PRM and improving the navigational path by smoothening the corners. In the study, the terms "PRM path planning", "probabilistic roadmap", "mobile robot path planning" and "path smoothening" have been scanned using ScienceDirect, IEEE Xplore, SCOPUS, Web of Science, and Google Scholar to specify references that could influence the conceptional framework meticulously. To ensure the comparability of the findings, only studies that have used mobile robot path plannings in the last ten years with smoothening techniques have been included. This research resulted in the following evaluation layout:
- <u>Aim and problem definition:</u> It determines the motivation for mobile robot path planning to be considered and included in the scope of application.
- <u>Path smoothening method:</u> It includes path-creating methods.
- <u>Evaluation criteria:</u> It includes which methods and techniques were examined for the application goal.

- Evaluation of results: It covers the level of shortening the path, path smoothing degree, and calculation times.

As a result of the literature review of these elements, 26 academic studies were analyzed detailly. The main goal here is to examine the suitability of smoothening every corner of a navigational mobile robot path after deriving the path from PRM. From this point, PP techniques and the path smoothing process have been analyzed, and gaps in the literature on PP have been mentioned at the end of the section. This comprehensive scope also shows the necessity and contributions of the proposed methodology.

When PP literature is examined, it is seen that there are many algorithms to derive a navigational path from a robot environment, such as Probabilistic Roadmap Method (PRM), Random Trees, Rapidly Exploring Random Trees (RRT), RRT*, PSO, GA, ABC, APF, MD and VD which have been studied for mobile robot path planning problems. However, PRM is the most common PP method for practical reasons mentioned in Section 0. The main difficulties for planning a path are path selection, computational complexity, adaptability, and avoiding obstacles. Obtaining the shortest and smoothest distance is the main goal for researches. PP is also relevant in some areas such as autonomous exploration, molecular biology, search and rescue, manufacturing, computer-assisted surgery other than robotics.

Obtaining the roadmap for PP requires many different steps even after getting the path; an improvisation may be necessary to make the path smooth and shorter. There is also a problem related to environment subdivision and characterize the subregions [15]. For dynamic configuration of spaces, the problem has also been tried to be solved by some methods as well Gaussian PRM Sampling [16], artificial potential method for node distribution [17], improved-based strategy [18], and potential fields by swarm robots [19]. On the other hand, sampling narrow passages is also a challenge for PRM planners. There are also some specific studies about narrow passages such as bridge test [20][21], sample-based [22]–[24] and sensor-based algorithms [25], node adding [26], A Star Algorithm [27].

Smoothening the sharp corners is also another work of PP problem. After obtaining the feasible path, researchers have been using some techniques for smoothing the path such as Bézier Curves [28]–[33], spline method [33]–[37], fuzzy control [9], [38], Genetic Algorithm (GA) [39]–[41], etc., which are summarized in Table 1. The common points of these academic studies are the final preparation of the navigational mobile robot path for smooth turnings.

**Table 1.** Literature summary on path smoothing methods

| Ref. | Publication Type* | Authors and year of publication | Path deriving method** | Path smoothing method** |
|---|---|---|---|---|
| [28] | JA | (Song et al., 2021) | Improved PSO | High degree Bézier Curve |
| [34] | JA | (Ayawli et al., 2021) | MD, VD and A* | Cubic spline interpolation |
| [42] | JA | (Aria, 2020) | PRM and RRT | Reed sheep planner |
| [35] | JA | (Sun et al., 2020) | VD and RRT-Connect | Cubic spline |
| [9] | JA | (Mohanta & Keshari, 2019) | PRM | Fuzzy control system |
| [37] | JA | (Song et al., 2019) | Modified PSO | η3-splines |
| [29] | JA | (Tharwat et al., 2019) | Chaotic PSO | Bézier curve |
| [39] | JA | (Nazarahari et al., 2019) | APF | Enhanced GA |
| [43] | CP | (Cheok et al., 2019) | - | Lyapunov stability |
| [30] | JA | (Elhoseny et al., 2018) | Modified GA | Bézier curve |
| [44] | CP | (Wang et al., 2017) | A partial and global A* | A partial and global A* |
| [36] | CP | (Sudhakara et al., 2017) | PRM | Spline method |
| [45] | CP | (Janjoš et al., 2017) | RRT | Quartic spline |
| [46] | JA | (Han & Seo, 2017) | Surrounding point set | Former and latter points |
| [8] | JA | (Gang & Wang, 2016) | PRM | Circular arc |
| [31] | JA | (Song et al., 2016) | GA | Bézier curve |
| [47] | JA | (Ravankar et al., 2016) | PRM | Hypocycloid curve |
| [32] | JA | (Simba et al., 2015) | A* | Bézier curve |
| [40] | JA | (Davoodi et al., 2015) | Multi objective GA | Multi objective GA |
| [48] | JA | (Contreras et al., 2015) | ABC | EPA |
| [38] | CP | (Su & Phan, 2014) | - | Fuzzy inference system |
| [49] | CP | (Huh & Chang, 2014) | PRM | Modified QPI |

| [33] | JA | (Yang et al., 2013) | - | Cubic-Bézier spiral |
| [50] | JA | (Kapitanyuk & Chepinsky, 2013) | - | Piecewise method |
| [51] | JA | (Yang et al., 2012) | Voronoi tessellation | SVM |
| [41] | CP | (Ju & Cheng, 2011) | GA | GA |

*Journal Article (JA); Conference Paper (CP).
**The full name of the methods has been presented in Nomenclature.

The path smoothing process in the literature mentioned in Table 1 has some disadvantages such as too much computation time or only working when the path has a few nodes or not considering path deriving method to be optimal, and many of paths would be smoother and shorter even after smoothing process is performed.

This paper proposes optimal smooth path planning, which takes the path from PRM in a complex environment to contribute to research in the field. It shows that some adjustments at PRM affect the performance of smoothening process on time complexity and for the shorter distance. Two steps were studied in this paper:
1) The first step is determining feasible connection distance between nodes at pure paths obtained from PRM.
2) The second step is smoothening every node at the path derived from PRM using our method, which is the arc fillet method.

The proposed path smoothing method is easy to perform, has minimum calculation time, and can be successfully applied to many applications like robot manipulator movement planning, mobile robot path planning, multiple robot path planning, intelligent unmanned aerial vehicle navigation, etc. As a mobile robot path construction method, PRM was chosen in this study for its advantages as a tool to derive a path.

## 4. The proposed Arc Fillet method

Smoothening process of the edges of a path consists of two steps explained below. The smoothening process is to be performed by helping a circular arc that belongs to a circle. The present research aims to find the right circle and position it in the right place. The purpose of smoothening here is not just for the sharp edges; it can be applied to all sorts of edges with angles varying from till ö180º.

### a. Finding the optimal cutting points

There mustn't be any remaining sharp edges after smoothening the corners at a given section of the path; this depends not only on the radius of the circle but also on the cutting points.

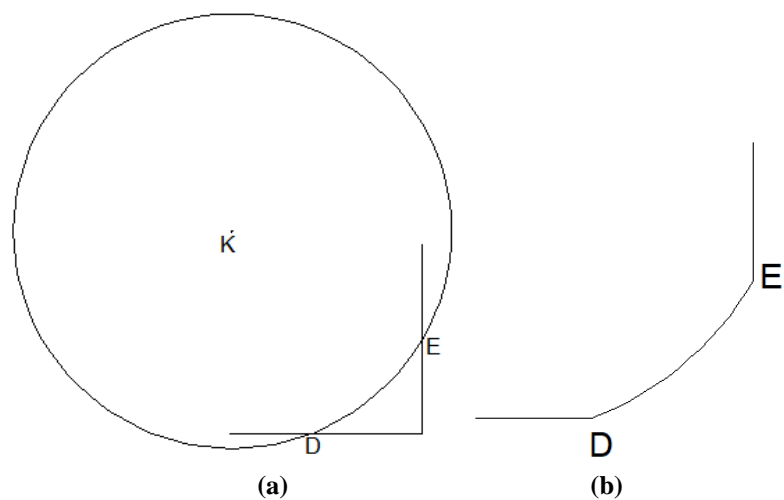

**Fig. 5.** A random circle cutting the edges of a path

In Fig. 5a, the K centered circle cuts the edges of a path from D and E points. After trimming the remaining part from the node side, the modified path can be seen in Fig. 5b. Unfortunately, the modified path doesn't seem to be smooth enough and still needs to be smoothened.

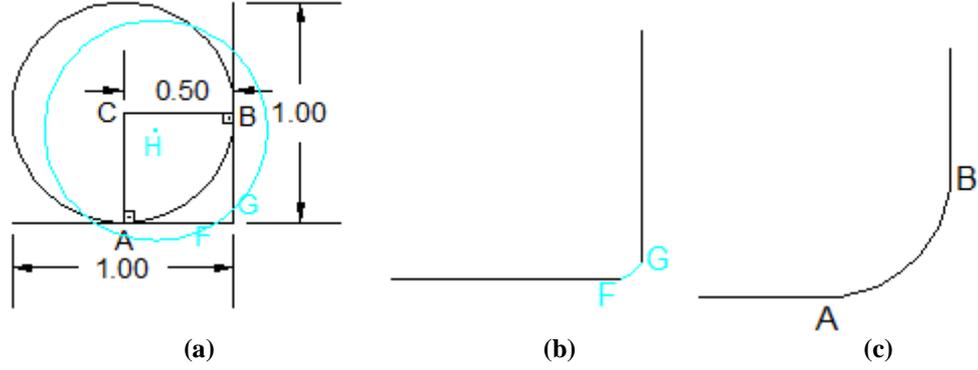

**Fig. 6.** Two same circles cutting the edges of a path from different points

In Fig. 6, there are two circles placed on a given edge of a path. Although they have the same diameter, cutting points make a difference in the smoothness of the new path. After trimming the remaining part from the corner side, two small corners appear at the F and G points in Fig. 6b. Only in Fig. 6c arc AB connects the edges without any interruptions; in other words, no different small corners appear after trimming.

The optimal circle whose arc is compatible with the edges for smoothness must tangentially touch the edges. The circle mustn't cut the edges to avoid having extra mini corners. For a path that has many nodes, the formulation of the distance formula between two nodes would be:

$$disth_i = \sqrt{(x_{i+1} - x_i)^2 + (y_{i+1} - y_i)^2} \tag{1}$$

At Eqn. (1), the distance between two nodes is calculated from node number $i$ to $m$, where m denotes the number of nodes at the path.

To find the points that smoothening arcs start and end with, first, we need to define which edge is shorter. Let's suppose these points are $a_i$ and $b_i$. Then let the ratio of distances is $t$, measured from $a_i$ and $b_i$ to the corner divide to the whole edge.

If $dist_i < dist_{i+1}$ half of the shorter edge ($h$) is $dist_i/2$. Then the ratio of distance;

$$t = h/2 \tag{2}$$

Starting point coordinates of smoothening arc at the shorter edge would be:

$$a_{xi} = (x_i + x_{i+1})/2 \tag{3}$$
$$a_{yi} = (y_i + y_{i+1})/2 \tag{4}$$

Ending point coordinates of smoothening arc at the longer edge would be:

$$b_{xi} = (1-t)x_{i+1} + tx_{i+2} \tag{5}$$
$$b_{yi} = (1-t)y_{i+1} + ty_{i+2} \tag{6}$$

If $dist_i >= dist_{i+1}$ half of the shorter edge ($h$) is $dist_{i+1}/2$. The ratio of distance is still the same equation in Eqn. (2).

Starting point coordinates of smoothening arc at the shorter edge would be:

$$a_{xi} = (1-t)x_{i+1} + tx_i \qquad (7)$$
$$a_{yi} = (1-t)y_{i+1} + ty_i \qquad (8)$$

Ending point coordinates of smoothening arc at the longer edge would be:

$$b_{xi} = (x_{i+1} + x_{i+2})/2 \qquad (9)$$
$$b_{yi} = (y_{i+1} + y_{i+2})/2 \qquad (10)$$

### *b. Finding optimal diameter*

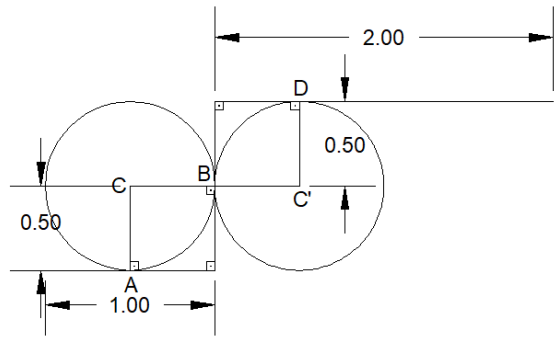

**Fig. 7.** Defining the circle diameter

In Fig. 7, the half of the edge with point B creates identical circles of C and C'. The common point of these circles is point B. When the edges have different lengths (the edge that has point B and the edge that has point D), the circle that smoothens the corner depends on the shorter edge. For maximum smoothness, the path continues by drawing an arc from A to B, and another arc starts at B point without stopping. For optimum smoothness, the circle's diameter equals to half the length of the shorter edge; this is also the same case for any edges whose angle varies.

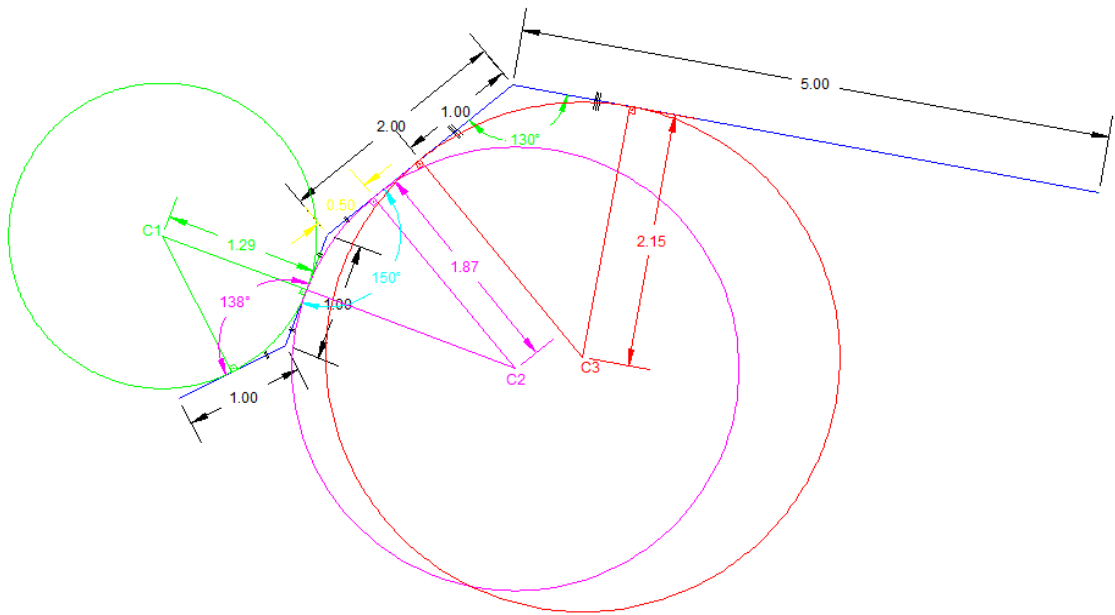

**Fig. 8** A path has different angles between edges

In Fig. 8, the edges of the path have three different lengths, and the angles between them are also different; this leads to different smoothening circles having different sizes. The edges that C1 centered circle tangent to, have the same length. One of the edges of these is also tangent to C2 centered circle. The size of the circle depends on the angle between the edges and the tangent points at the edges.

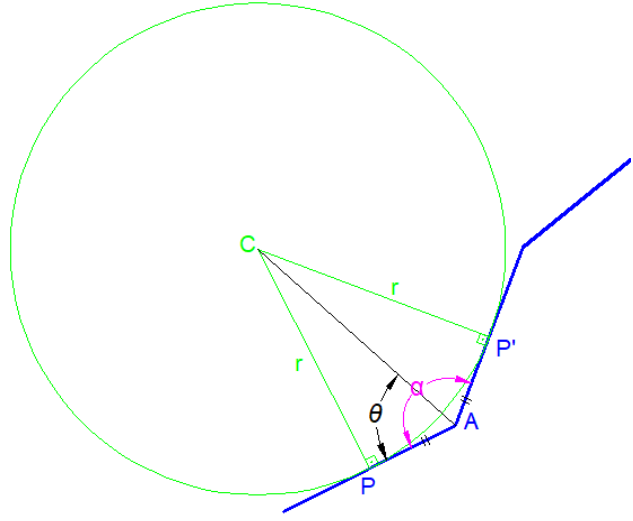

**Fig. 9** Finding the circle diameter

In Fig. 9, the triangle of CAP and CAP' are identical because circle C is tangent to the path's edges. CP and CP' are the diameter of this circle. The angle of CAP and CAP' are the same because of identical triangles. Hence, these angles ($\theta$) are half of the angle between edges ($\alpha$). The length of AP and AP' is the half of the shorter edge. To find the radius of the smoothening circle first, we need to find the angle between edges by:

$$\alpha_i = \text{Arctan}(\frac{y_i - y_{i+1}}{x_i - x_{i+1}}) - \text{Arctan}(\frac{y_{i+2} - y_{i+1}}{x_{i+2} - x_{i+1}}) \tag{11}$$

At Eqn. (11) $\theta_i$ is the angle between $i_{th}$ and $(i+1)_{th}$ edges where coordinates at $i_{th}$ nodes which are $(x_i, y_i)$ shaped. If the angle is negative, then;

$$\alpha_{inew} = 2\Pi\alpha_i \tag{12}$$

Finding radius is required to find the angle between edges and the vector from the circle's center to the node. At $i_{th}$ node, this angle would be:

$$\theta_i = \frac{\alpha_i}{2} \tag{13}$$

With the help of simple trigonometric equations, the radius of the smoothening circle between edges can be found by:

$$r_i = \tan(\alpha_i).|AP| \tag{14}$$

At Eqn. (14), $|AP|$ can be seen in Fig. 9 as the half of the shorter edge.

From Fig. 9, the center coordinates of the center can be found by Eqn. (15). Here, the C point is the circle's center, and P is the tangential touch to the one edge.

$$(P_x - C_x)^2 + (P_y - C_y)^2 = r^2 \qquad (15)$$

This circle also tangentially touches the next edge. Therefore, using the same equation for the next point reduces the number of circles that touch these two points.

$$(P'_x - C_x)^2 + (P'_y - C_y)^2 = r^2 \qquad (16)$$

From Fig. 9, the center coordinates of the circle can be found by Eqn. (16) too. Here, the C point is the circle' center, and P' is the tangential touch to the next edge.

To this point, we reduced the number of circles that touch defined points at the edges to two. These two circles are identical, have the same radius, and connect the edges to the same two points tangential, but these circles have a different direction seen in Fig. 10.

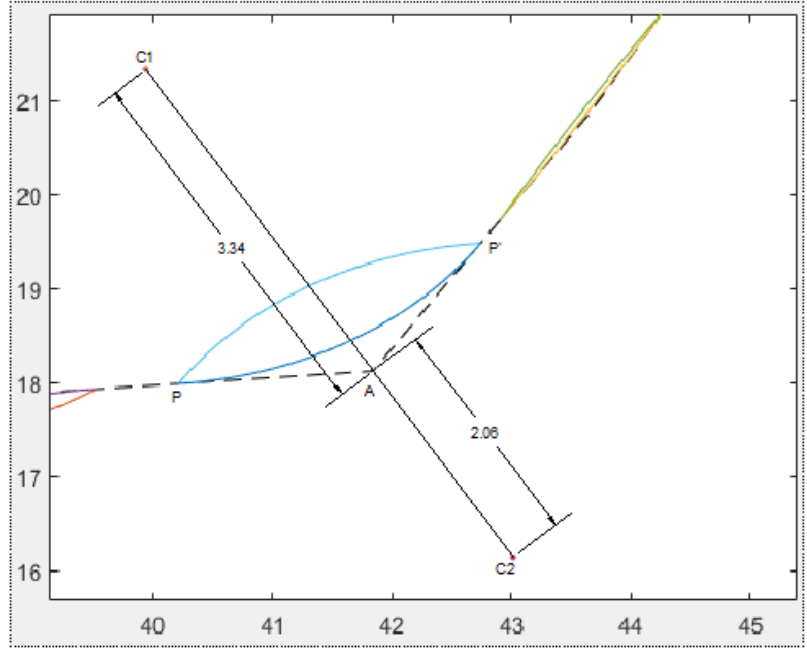

**Fig. 10** Finding the correct arc

In Fig. 10, C1 and C2 centered arcs touch the same path edges to the same points (Point P and P'). The correct arc that smoothens the path has the center that has more distance from corner A than the incorrect arc. Eqn. (15) and (16) give these double center coordinates of these arcs. Let's suppose these coordinates are $C1_x$ and $C1_y$ for the C1 centered arc and $C2_x$ and $C2_y$ for the C2 centered arc. And we named the correct arc's center as C. Then:

$$C_i = \begin{cases} C2_i & \text{if } \sqrt{(x_i - C1_{ix})^2 + (y_i - C1_{iy})^2} < \sqrt{(x_i - C2_{ix})^2 + (y_i - C2_{iy})^2} \\ C1_i & \text{if } \sqrt{(x_i - C1_{ix})^2 + (y_i - C1_{iy})^2} > \sqrt{(x_i - C2_{ix})^2 + (y_i - C2_{iy})^2} \end{cases} \qquad (17)$$

At Eqn. (17), in a path, $i_{th}$ numbered node's coordinates were named as $x_i$ and $y_i$. The correct arc that smoothens the corners was assigned as $C_i$ at node number $i$.

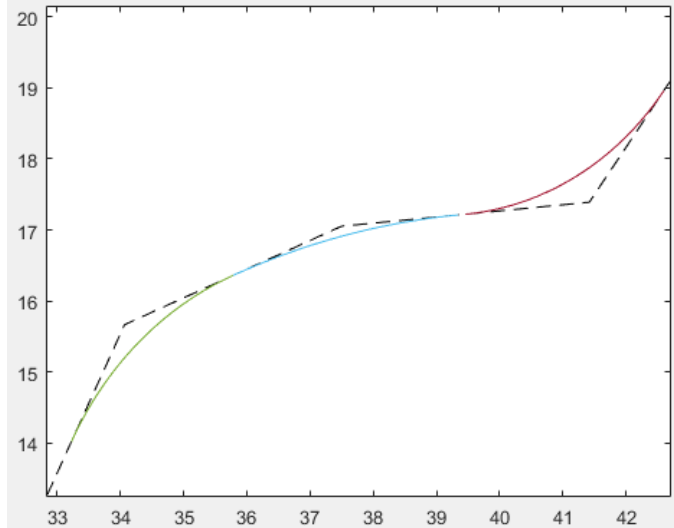

**Fig. 11** Final smoothening arcs of a path

After processing corners for smoothening in a path, results can be seen successfully in Fig. 11. The process from circles to arc and the final path can be seen in Fig. 12 and Fig. 13.

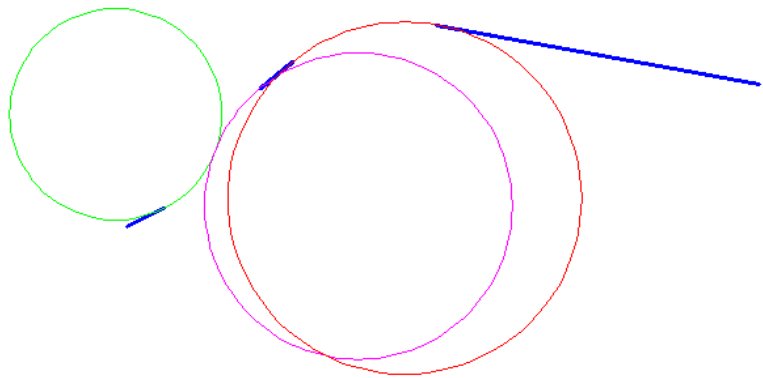

**Fig. 12.** Trimming the path with the optimized circles

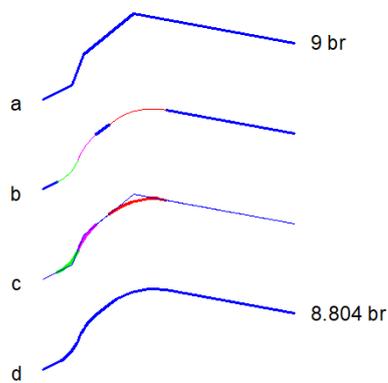

**Fig. 13.** Remainder after smoothening the edges

The suitable circles in Fig. 12 trimmed the corners at the path in Fig. 8. After lifting out the circles from the corners, the smoothened path can be seen in Fig. 13d. By smoothening the path, the path length is also diminished. The total length of the whole path can be found by:

$$dist = \sum_{i=1}^{m} \sqrt{(x_{i+1} - x_i)^2 + (y_{i+1} - y_i)^2} \quad (18)$$

At Eqn. (18), the distance of the path is calculated by the *dist* equation from node number *i* to *m*, where *m* denotes the number of nodes at the path.

The path planning scheme is explained with the help of the flowchart shown in Fig. 14.

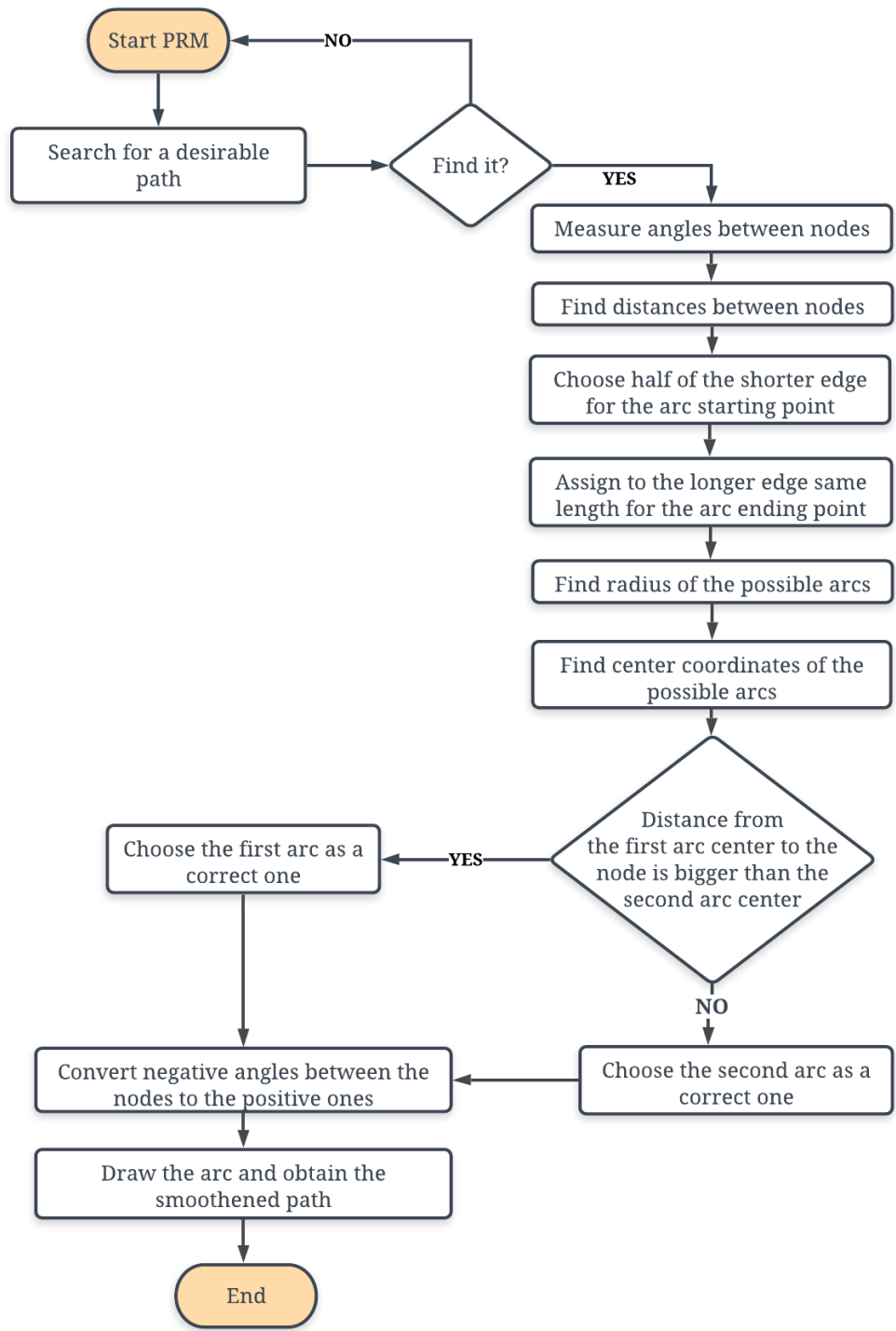

**Fig. 14** Flow diagram defining the PRM smoothening path

## 5. Case study

**Workspace preparation:** The robot's safety may not be guaranteed if its dimension is not considered while building the path. The path may be built close to obstacles or could also be built in narrow passages more diminutive than the robot's size, resulting in a collision. Besides, the robot may not use the path fully as formed because of noise and other uncertainties caused by hardware restrictions and other systematic and non-systematic flaws. There may be deviations resulting in collisions. Using PRM to generate a path does not consider the robot's size, which may cause crashes and trapping of the robot in narrow passages. While addressing this, and taking care of uncertainties, e.g., obstacles were inflated based on safety space requirements as mentioned in Section 0.

Many simulations have been tested on a standard complex map at MATLAB (binary occupancy map) for the same starting and endpoints. At a different number of PRM nodes and different connection distances, both pure PRM path and smoothened path behaviors have been observed. For upper and lower limits of connection distance, map and obstacle measurement limits have also been considered.

After PRM shaped the path, we smoothened the path, and we made a simulation robot for walking the smoothened path from beginning to end. We observed some outcomes in the coming sections. Map, simulation robot, and controller properties at MATLAB are given in Table 2, Table 3, Table 4, and Table 5.

**Table 2.** Map

| | |
|---|---|
| Type | Binary occupancy |
| Data Type | Logical |
| X limits (meters) | [0 52] |
| Y limits (meters) | [0 41] |
| Resolution | 1 |

**Table 3.** Controller

| | |
|---|---|
| Type | Pure pursuit |
| Max. Angular Velocity | 2 |
| Look Ahead Distance | 0.3 |
| Desired Linear Velocity | 0.6 |

**Table 4.** Robot

| | |
|---|---|
| Type | Differential drive |
| Track Width | 1 |
| Sample Time | 0.1 |
| Frame Size | 1.25 |

**Table 5.** Viz rate

| | |
|---|---|
| Type | Rate control |
| Desired rate | 10 |
| Desired period | 0.1 |

### a. *Simulation environment-1*

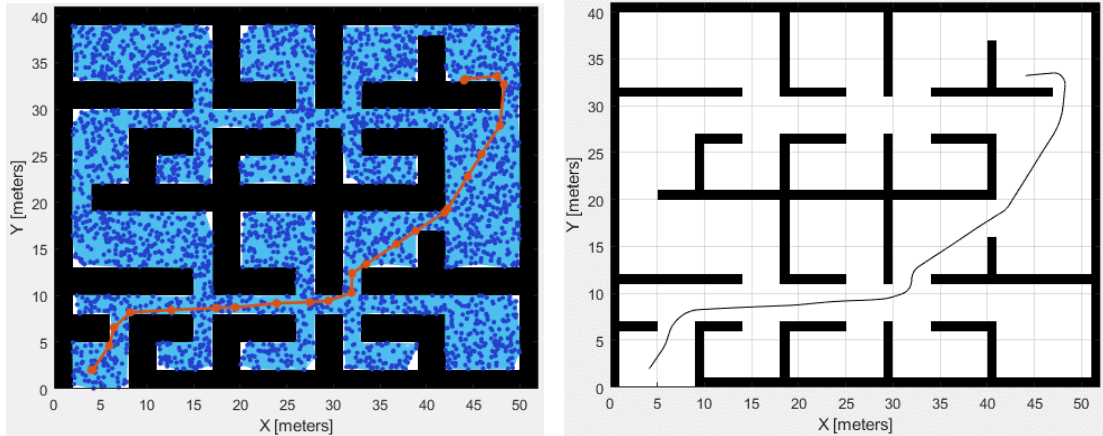

**Fig. 15. (a)** PRM by inflated map, number of nodes:3000, connection distance:5
**(b)** Smoothened path of **(a)**.

In Fig. 15a, PRM worked on an inflated map. If an inflated map shapes the path, changing coordinates won't affect the obstacles when shaping the path for smoothening. Initially, at PRM, 3000 nodes are shot to the obstacle-free area. The Connection distance was chosen 5 meters. It took 7,15 seconds to shape the path. The path length is 65.97 meters, and the path has 25 nodes on it. In Fig. 15b, to smoothen the path, 0.54 seconds have passed. The length of the smoothened path is 64.95 meters, and the simulation robot walked by this smoothened path from start to end within 112.15 seconds.

### b. Simulation environment-2

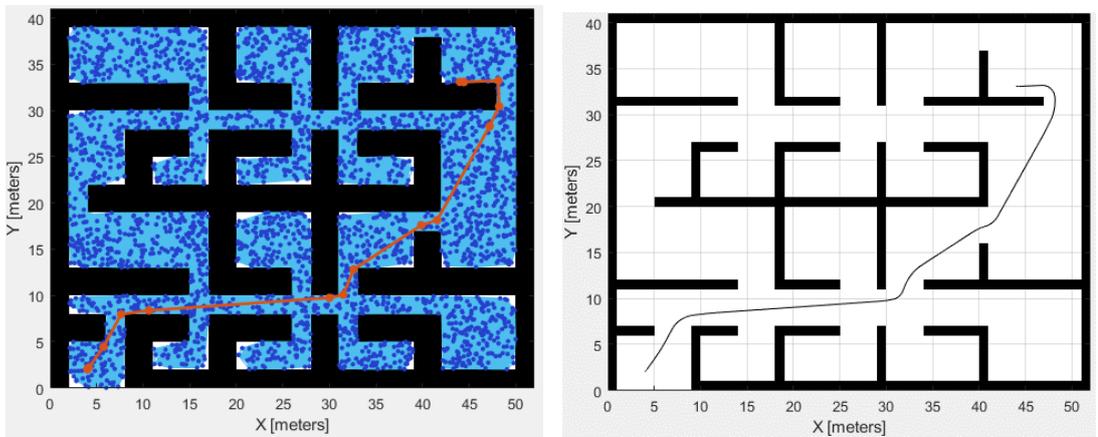

**Fig. 16. (a)** PRM by inflated map, number of nodes:3000, connection distance:100

In Fig. 16a, PRM worked on an inflated map. Initially, at PRM, 3000 nodes are shot to the obstacle-free area. The connection distance was chosen 100 meters, and it took 140.36 seconds to shape the path. The path length is 140.36 meters, and the path has 15 nodes on it. In Fig. 16b, to smoothen the path, 1.03 seconds have passed. The length of the smoothened path is 64.14 meters, and the simulation robot walked by this smoothened path from start to end within 107.38 seconds.

### c. *Simulation environment-3*

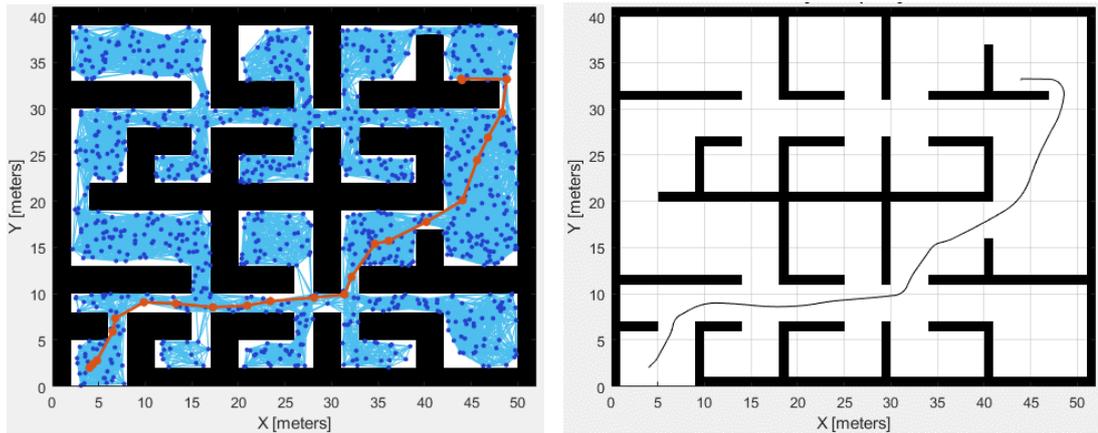

**Fig. 17. (a)** PRM by inflated map, number of nodes:1000, connection distance:5
**(b)** Smoothened path of **(a)**

In Fig. 17a, PRM worked on an inflated map. Initially, at PRM, 1000 nodes are shot to the obstacle-free area. The connection distance was chosen 5 meters, and it took 9.18 seconds to shape the path. The path length is 67.46 meters. The path has 23 nodes on it. In Fig. 17b, to smoothen the path, 6.69 seconds have passed. The length of the smoothened path is 65.74 meters, and the simulation robot walked by this smoothened path from start to end within 110.74 seconds.

### d. *Simulation environment-4*

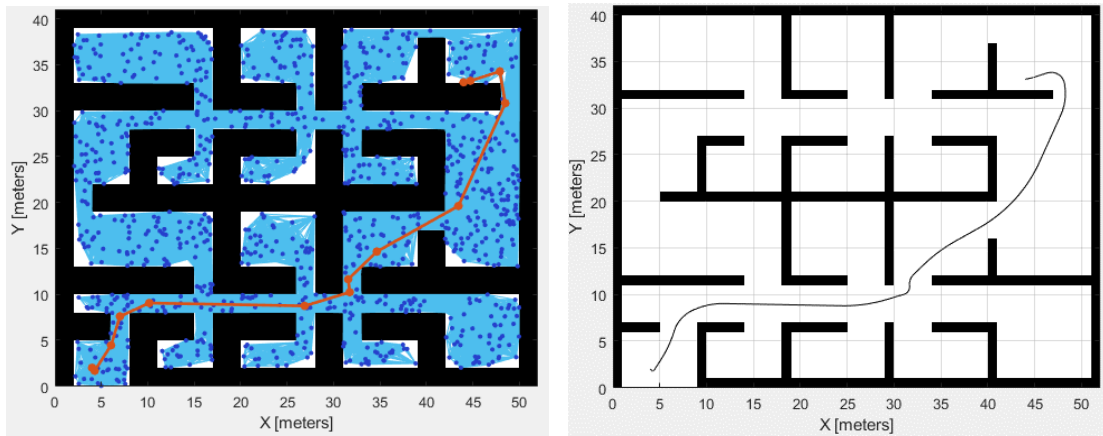

**Fig. 18. (a)** PRM by inflated map, number of nodes:1000, connection distance:100
**(b)** Smoothened path of **(a)**.

In Fig. 18a, PRM worked on an inflated map. Initially, at PRM, 1000 nodes are shot to the obstacle-free area. The connection distance was chosen 100 meters, and it took 14.55 seconds to shape the path. The path length is 68.00 meters, and the path has 14 nodes on it. In Fig. 18b, to smoothen the path, 0.93 seconds have passed. The length of the smoothened path is 65.89 meters, and the simulation robot walked by this smoothened path from start to end within 109.81 seconds.

The properties of these figures can be seen in Table 6.

**Table 6.** Comparison of paths

| | Fig. 15 | Fig. 16 | Av.of 20 Samp. | Av.of 20 Samp. | Fig. 17 | Fig. 18 | Av.of 20 Samp. | Av.of 20 Samp. |
| --- | --- | --- | --- | --- | --- | --- | --- | --- |

|  | PRM Nodes | 3000 | 3000 | 3000 | 3000 | 1000 | 1000 | 1000 | 1000 |
|---|---|---|---|---|---|---|---|---|---|
|  | Conn. Dist. | 5 | 100 | 5 | 100 | 5 | 100 | 5 | 100 |
| PRM | Time | 7.15 | 140.36 | 38.12 | 171.35 | 9.18 | 14.55 | 2.36 | 14.86 |
|  | Path Length | 65.97 | 65.33 | 66.32 | 66.13 | 67.46 | 68.01 | 68.60 | 67.65 |
|  | Path Nodes | 25.00 | 15.00 | 23.30 | 14.35 | 23.00 | 14.00 | 23.00 | 13.35 |
| Smooth Path | Path Length | 64.95 | 64.14 | 65.11 | 64.95 | 65.74 | 65.89 | 66.69 | 65.71 |
|  | Time | 0.54 | 1.03 | 1.82 | 1.06 | 6.69 | 0.93 | 1.77 | 1.00 |
|  | Robot Walking Time | 112.15 | 107.38 | 109.70 | 108.66 | 110.74 | 109.81 | 111.92 | 111.12 |
| Total Time |  | 119.85 | 248.77 | 149.64 | 281.07 | 126.60 | 125.29 | 116.05 | 126.98 |

In Table 6, both pure path by PRM and smoothened version have been observed at different PRM node numbers and connection distances to shape the path. 3000 and 1000 PRM Nodes shaped the path at the same connection distance. Paths have also been observed at a maximum connection distance of 5 meters and 100 meters to shape the path. In Table 6, the properties of the paths can be seen in figures in the Simulation Environments sections. Twenty samples of each category have also been simulated to observe the average behaviors of the same variables.

## 6. Results and discussion

This section has made comparisons among the different PRM nodes at CS and connection distances between nodes while shaping the path. Path smoothing performances have been observed during four case studies in simulation experiments. When connecting the nodes to shape the path, the nearest possible nodes are connected, or the furthest possible nodes are connected. Even these connection distances, while shaping the path in the first place, also affects the behaviors of both the pure path and the smoothened one. The PP performance was mainly evaluated on path lengths, the number of path nodes or connection distance between nodes, and time is taken to cover the trajectories.

### a. Number of PRM nodes

While shaping the path by PRM, the number of nodes at the movement area affects the path shape and time complexity.

For the pure path shaped by PRM:
- At the same connection distance, while PRM nodes are increasing, the time for shaping the path out of them also increases.
- If there are more nodes at CS, then the path has more segments.
- Pure path length shaped by PRM is shorter at more nodes at the CS.

For the smoothened path:
- When more PRM nodes shape the path, then the smoothened path is shorter.
- Time taking smoothening the path is more when more PRM nodes shape the path.
- When a simulation robot walks at the smoothened path, it takes less time to finish it when more PRM nodes shape the path.
- Total time of shaping PRM, smoothening the path, and the robot walking on it is more when more PRM nodes shape the path.

### b. Connection distance between PRM nodes

While shaping the path by PRM, the connection distance of the nearest node at the path affects the path shape and time complexity.

For the pure path shaped by PRM:
- When the connection distance between the nodes is more while shaping the path, the time needed for shaping the path is also more.
- The path length is more when the connection distance is less.

- Nodes at the path are more when the connection distance is less.

For the smoothened path:
- The path length is more when the connection distance is less.
- Time taking smoothening the path is more when the connection distance is less.
- When the simulation robot walks at the smoothened path, it takes less time to finish it when the connection distance is more.
- Total time of shaping PRM, smoothening the path, and the robot walking on it is more when the connection distance is greater.

### 7. Conclusion and future work

This research proposes a new easy and practical circular arc fillet methodology that smooths and the navigation along the PRM path, makes some adjustments to derive optimal path from PRM. The proposed methodology defeats many drawbacks of PRM based navigational mobile robot trajectory, i.e., PRM obtained mobile robot paths are segmented in linear lines, and connecting of two linear lines generates the corner at the PRM path nodes; causes sharp turning risks, and limits hassle-free controlled velocity navigation for robots; provides an optimal shortest path between the start point and the goal point. After the path is derived from PRM, optimal circles are found to smoothen the corners. Circular arc parameters such as radius and circle intersections with the edges need to be calculated for optimal circles. Before constructing the path, obstacles are inflated to have a safe environment after smoothing the path. The smoothing methodology is performed on four different PRM features. The modified circular arc method outperforms to handle any problematic sharp turnings in obstacle present environment.

It is mentioned in [9] that drawing a circular arc can be difficult if path joints are very close to the obstacles, and determining a suitable radius of the circular arcs is itself a difficult job. However, this study demonstrates that determining the appropriate radius of the circular arcs becomes easy to compute and perform with the new geometric approach proposed in this paper. Furthermore, smoothened path safety is ensured by inflating obstacles before applying PRM.

Smoothening the path is affected by some factors, even different PRM variables initially while shaping the pure path. Results show that when the path has fewer nodes, in other words, when the connection distance between the nodes is more, the path is straighter; as a result, the path is shorter. When the path has more nodes and less connection distance, it requires additional work to smoothen extra nodes. However, shaping the original pure path by PRM at the beginning with more connection distance requires more time.

For future work, before smoothing the path, PRM path construction time with more connection distance can be reduced by processing the path with some techniques. After improving the pure path, the smoothing process can be applied to a straighter path. This study has not been tested in a dynamic environment, and maybe it would be considered in future works.

At this work, PRM algorithm was used to obtain a mobile robot navigational path. Circular arc fillets eliminated path corners. Finally, the path became ready to have smooth and shorter navigation. This smoothening process can be applied to any PP algorithm that gives path coordinates as an outcome and can be successfully performed in real life mobile robot PP applications such as autonomous robot navigation PP, multiple robot PP, robot manipulator movement planning, intelligent unmanned aerial vehicle navigation, etc.